\documentclass{article}

\PassOptionsToPackage{numbers,sort&compress}{natbib}
\usepackage[preprint]{neurips_2026}
\usepackage{graphicx}

\usepackage[utf8]{inputenc} % allow utf-8 input
\usepackage[T1]{fontenc}    % use 8-bit T1 fonts
\usepackage{hyperref}       % hyperlinks
\usepackage{url}            % simple URL typesetting
\usepackage{booktabs}       % professional-quality tables
\usepackage{multirow}
\usepackage{amsfonts}       % blackboard math symbols
\usepackage{nicefrac}       % compact symbols for 1/2, etc.
\usepackage{microtype}      % microtypography
\usepackage{xcolor}         % colors

\newcommand{\R}{\mathbb{R}}
\newcommand{\methodname}{PhotoFlow}
\newcommand{\benchmarkname}{VPhotoBench}

\title{\methodname: Agentic 3D Virtual Photography Missions}

\author{\normalfont
Jiarui Guo$^{1,2}$ \quad
Haojia Wei$^{3}$ \quad
Yiming Zhang$^{4}$ \quad
Yifei Liu$^{5}$ \quad
Yuning Gong$^{6}$ \\
Hongjie Zhang$^{5}$ \quad
Xue Yang$^{1}$ \quad
Zhihang Zhong$^{1}$\thanks{Corresponding author.  Email: zhongzhihang@sjtu.edu.cn} \\
{\small $^{1}$Shanghai Jiao Tong University \quad $^{2}$Northeastern University \quad $^{3}$University of California, Los Angeles} \\
{\small $^{4}$Cornell University \quad $^{5}$Shanghai AI Laboratory \quad $^{6}$Sichuan University} \\
{\small \url{https://visionary-laboratory.github.io/PhotoFlow/}}
}

\begin{document}

\maketitle

\begin{abstract}
Virtual photography asks an agent to enter a prepared 3D scene with no preselected camera pose or reference image, infer a suitable shot from scene information and a language intent, choose executable camera parameters, and render the final photograph. Recent progress in vision-language models makes this kind of spatial agent increasingly plausible, but the task stresses two capabilities that remain hard to evaluate together: complex 3D spatial understanding and abstract aesthetic judgment. We introduce \methodname, a Director-Reviewer-Reflector agent for closed-loop camera search. The Director builds a soft photographic blueprint and proposes diverse candidate cameras; the Reviewer combines rule checks, visual critique, and pairwise incumbent selection; and the Reflector converts failures into region memory, dead-zone suppression, and high-explore relocation. We also introduce \benchmarkname, a benchmark of 47 open-license Blender scenes and 141 language-conditioned photography missions spanning subject placement, relational composition, and atmosphere/style. On held-out experiments, \methodname{} achieves the strongest external quality-alignment composite and success rate among one-shot prediction, single-chain reflection, anchor-bank selection, and random search under a six-round rendering budget. To our knowledge, this is the first work to make language-conditioned virtual photography in arbitrary Blender scenes an executable agent task, and our results show that an LLM-centered spatial agent can already produce strong photographs in a setting designed to challenge both 3D reasoning and aesthetic choice.
\end{abstract}

\begin{figure}[t]
  \centering
  \includegraphics[width=\linewidth]{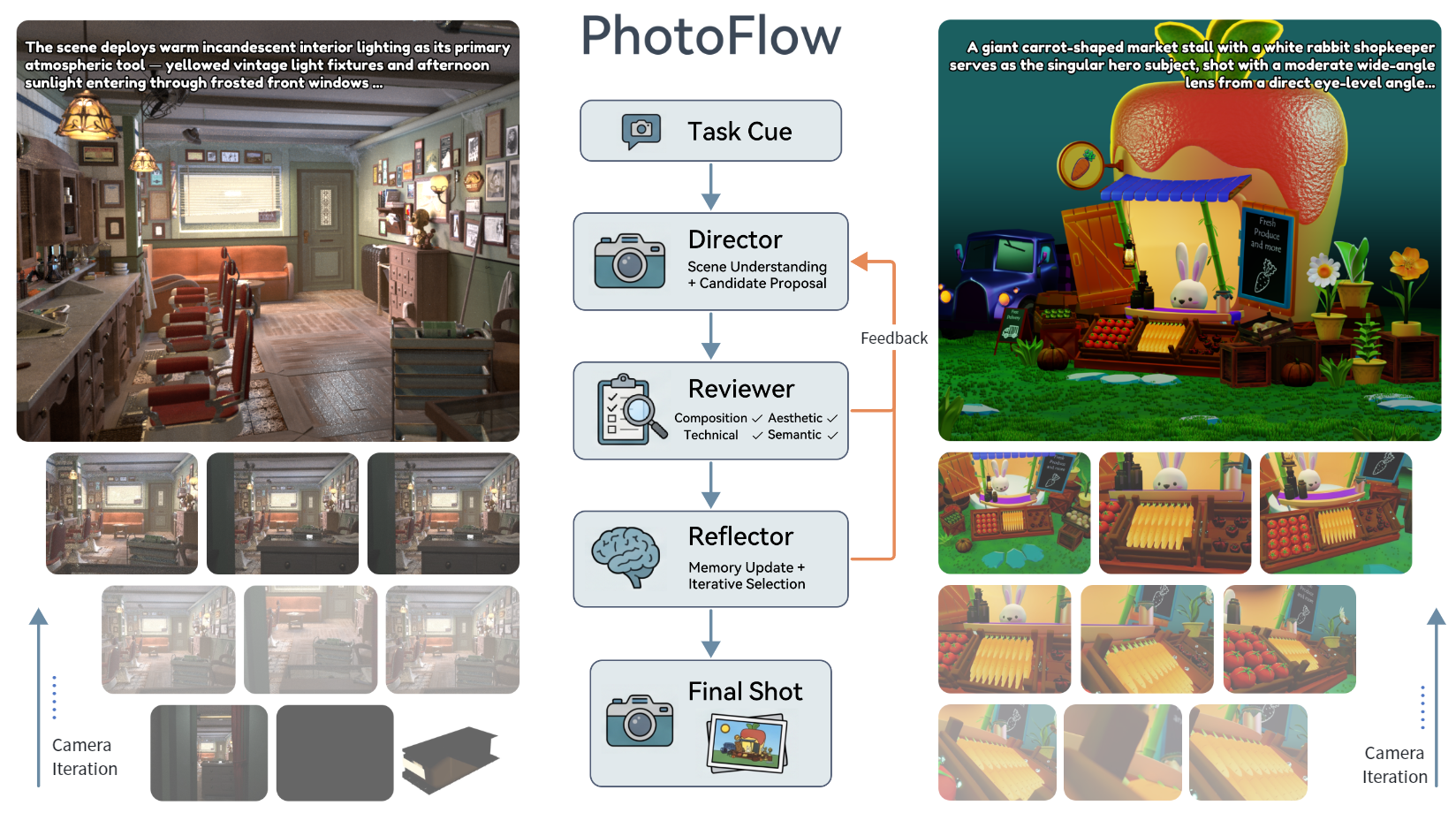}
  \caption{\textbf{Virtual photography as spatial-aesthetic decision making.} Given a controllable 3D scene and a language instruction, the agent must choose an executable camera state that satisfies spatial constraints, semantic intent, and photographic quality. The benchmark evaluates the final rendered image together with the search process that produced it.}
  \label{fig:teaser}
\end{figure}

\section{Introduction}

Virtual photography builds on automated camera control and virtual camera planning, where a system must choose concrete camera specifications to communicate a scene through composition and viewpoint \citep{he1996virtual,christie2005survey,abdullah2011composition,byers2004robot,alzayer2021autophoto}. We study the language-conditioned version: given a controllable 3D scene and a photography intent, a spatial agent must produce a final still image by choosing an executable camera state. Unlike image generation, the output camera pose, look-at target, lens, aperture, and aspect ratio must correspond to a rerenderable view of the scene. The task therefore joins two requirements that are usually evaluated separately: the agent must understand 3D layout and visibility, and the rendered image must satisfy an abstract photographic goal such as subject emphasis, relational composition, or atmosphere.

This combination exposes a difficult gap in current multimodal intelligence. Vision-language models remain unreliable on spatial relations, object orientation, relative depth, and multi-view perception, even in controlled benchmarks with visible objects \citep{liu2023vsr,kamath2023whatsup,fu2024blink,wang2024picture,stogiannidis2025mind}. Aesthetic evaluation is also not a settled oracle: image-aesthetic and perceptual-quality models are useful proxies, but human preference is subjective and depends on both image attributes and viewer factors \citep{murray2012ava,talebi2018nima,yang2022para,cao2025unipercept}. Virtual photography stresses both sides at once because the agent must search through physically valid 3D views while optimizing for a high-level visual intent.

No existing benchmark directly covers this setting. Robotic photography emphasizes physical capture, drone cinematography emphasizes smooth trajectories, aesthetic assessment scores completed images, embodied navigation evaluates paths, and text-to-image generation need not produce a valid camera state. To our knowledge, this is the first work to study language-conditioned still photography in arbitrary virtual art scenes as an executable agent task. Because no established public baseline suite exists for this exact problem, we construct controlled baselines that test one-shot prediction, single-chain reflection, anchor-bank selection, and random search, then use them to identify which failures appear and which agentic mechanisms mitigate them.

We introduce \methodname{}, a Director-Reviewer-Reflector agent that treats photography as finite-horizon feedback-driven search (Figure~\ref{fig:teaser}). The Director proposes diverse candidate cameras from scene scouts, a soft photographic blueprint, global anchors, and region memory; the Reviewer diagnoses rendered previews with rule-based and visual criteria; and the Reflector converts failures into search bias, dead-region suppression, and high-exploration relocation. We also introduce \benchmarkname{}, a 141-mission benchmark over 47 open-license Blender scenes. Under a six-round rendering budget, \methodname{} achieves the strongest external quality-alignment composite and success rate among the tested baselines, while the experiments report render-availability filtering, ablations, search diagnostics, and human consistency checks.

Our contributions are:
\begin{itemize}
  \item We propose \methodname, a Director-Reviewer-Reflector architecture for continuous camera search with soft blueprints, global anchor banks, region memory, four-dimensional review, pairwise incumbent selection, dead-zone suppression, forced high-explore relocation, and explicit aspect-ratio reasoning.
  \item We define \benchmarkname, a 141-mission benchmark over 47 open-license Blender scenes that couples scene geometry, natural-language intent, aspect-ratio choices, bootstrap protocols, and structured evaluation constraints.
  \item We report a held-out comparison with failure accounting, ablations, search diagnostics, human preference checks, and process analyses, so that final claims are tied to external metrics rather than internal reviewer scores alone. We will release the agent code, benchmark registry, task specifications, scene/license metadata, and evaluation scripts at \url{https://github.com/Visionary-Laboratory/PhotoFlow}.
\end{itemize}

\section{Related Work}

\paragraph{Automated photography and cinematography.}
Early automated photography systems treated camera placement as motion control under compositional constraints. The robot photographer of \citet{byers2004robot}, LeRoP \citep{kang2019lerop}, and reinforcement-learning methods such as AutoPhoto \citep{alzayer2021autophoto} demonstrate that camera placement can be automated as search. Drone and virtual cinematography systems further optimize subject tracking, smoothness, safety, and shot composition under real-time control constraints \citep{nageli2017realtime,bonatti2020autonomous,cinempc2024}. Language-driven systems such as ChatCam and recent film agents broaden the interface to conversational control, script-level planning, or multi-agent previsualization \citep{chatcam2024,filmagents2025,agenticAerial2025,minddirector2026}. Our task inherits the need for executable camera states, but differs in its design target: we study still photographic decision making in arbitrary-complexity virtual 3D scenes, where the final image must satisfy language-conditioned subject, relation, style, and aspect-ratio constraints rather than only reach a physical capture pose or produce a smooth trajectory.

\paragraph{Aesthetic assessment and view suggestion.}
Image aesthetic assessment provides the scoring tools that make automated photography measurable. Classic work studied photographic quality attributes and aesthetic datasets \citep{datta2006aesthetics,murray2012ava}; neural methods such as NIMA predict human aesthetic ratings from images \citep{talebi2018nima}; and Creatism demonstrated an end-to-end deep-learning photographer for professional-style image crops and post-processing \citep{creatism2017}. Recent 3D aesthetic-field approaches extend aesthetic prediction into continuous 3D viewpoint spaces \citep{tang2026aestheticfield}. These systems are important evaluators or priors, but they do not by themselves define a language-conditioned closed-loop agent that must reason about task constraints, aspect ratio, and iterative failures.

\paragraph{Embodied and virtual-environment benchmarks.}
Embodied AI benchmarks such as Matterport3D, Gibson, Habitat, and Room-to-Room have made navigation and spatial reasoning reproducible in 3D environments \citep{chang2017matterport,xia2018gibson,savva2019habitat,anderson2018r2r}. Their evaluation protocols make movement part of the task: navigation work commonly reports success together with path length or SPL \citep{anderson2018evaluation}, and VLN path-fidelity metrics such as nDTW and SDTW explicitly reward trajectories that follow the reference route \citep{jain2019stay}. LLM-based VLN agents such as NavGPT inherit this formulation by reasoning over navigation history and future explorable directions before choosing the next movement action \citep{zhou2023navgpt}. Virtual photography borrows the reproducibility discipline of embodied benchmarks, but it evaluates a different object: the final camera state and rendered image, not the route by which that state was discovered.

\section{\methodname}

\begin{figure}[t]
  \centering
  \includegraphics[width=\linewidth]{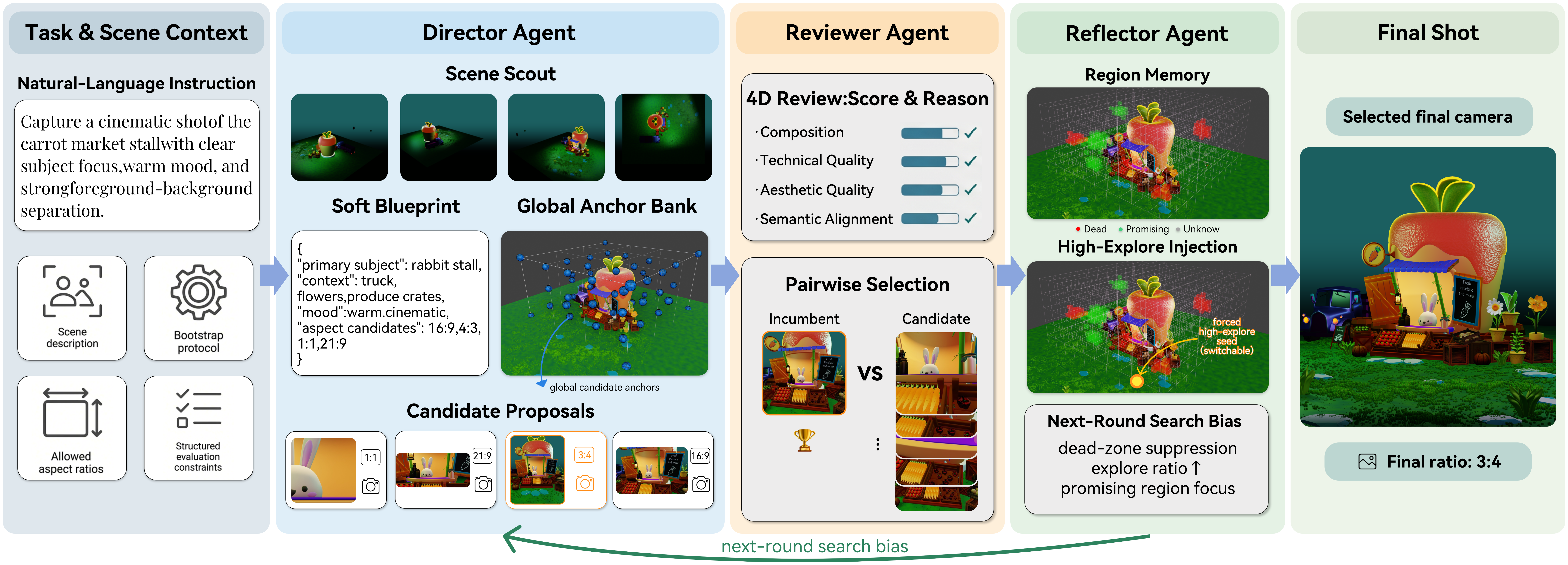}
  \caption{\textbf{\methodname{} pipeline.} The system first scouts the scene and constructs a soft photographic blueprint. The Director proposes candidate cameras from global anchors, region-memory-guided seeds, and a forced high-explore lane. Candidate previews are rendered in parallel, scored by a structured Reviewer, and summarized by a Reflector that updates search bias and forbidden regions for the next round.}
  \label{fig:method}
\end{figure}

\subsection{Task formulation}
\label{sec:task_formulation}

We define a virtual photography mission as a five-tuple
\begin{equation}
  b = (S, x, u, A, E),
  \label{eq:task}
\end{equation}
where $S$ is a controllable Blender scene, $x$ is a natural-language photography instruction, $u$ is the bootstrap information available to the agent, $A$ is the allowed aspect-ratio set, and $E$ is a structured evaluation specification. The specification $E$ is not a restatement of the prompt. It encodes checkable task intent such as primary subject visibility, screen placement, desired subject scale, camera-angle preference, symmetry, depth emphasis, and hard-failure conditions.

The output is an executable camera state
\begin{equation}
  c = (p, \ell, f, d, r),
  \label{eq:camera}
\end{equation}
where $p \in \R^3$ is the camera position, $\ell \in \R^3$ is the look-at point, $f$ is focal length, $d$ is aperture, and $r \in A$ is the selected aspect ratio. A renderer maps $(S,c)$ to an image $I=\mathcal{R}(S,c)$. This is the key difference from image generation: the final photograph must correspond to a concrete, rerenderable view of the scene. \methodname{} therefore does not directly regress $x$ to a single $c$; it performs finite-horizon search over $T$ rounds, rendering candidate views, receiving feedback, and updating its search bias.

\subsection{Scouting and blueprint}

Directly asking a large model to output continuous camera parameters from a raw object list is unstable. \methodname{} therefore begins with scene scouting. From Blender, it extracts three kinds of input. The geometric scene summary contains object names, bounding boxes, centers, scene extents, and coarse visibility proxies. The textual topology summary converts these statistics into relations such as dominant objects, foreground/background groups, vertical structure, and likely open regions. The global scout views are low-sample preview renders from a small set of canonical or visibility-oriented cameras around the scene. These observations give the language model explicit objects, coarse spatial relations, and visual anchors for relocation. The extracted scene blueprint is used as a photographic search substrate rather than a pedestrian reachability graph: in virtual production, a visually meaningful camera can be valid even when the set has no realistic entrance or traversable route to that position.

The Director then converts the instruction and scouting evidence into a soft blueprint. This conversion is an LLM parsing step with a constrained schema: the model identifies the likely primary subject, useful context objects, preferred composition cues, camera-angle preference, camera-zone preference, look-toward target, axis preference, symmetry preference, semantic vibe, and negative preferences. For example, an instruction asking for a ``lonely cinematic cabin'' may map to a small subject scale, a wider environmental frame, low or eye-level camera angle, and a muted semantic vibe. The blueprint is soft because these fields are preferences, not hard constraints: they bias search while allowing multiple valid photographs instead of forcing one template.

\subsection{Director}

The Director proposes candidates on top of interpretable spatial priors. A global anchor bank is a finite set of coarse camera seeds $\{a_i\}$ defined before local search begins. Each anchor contains an initial camera position, look-at target, approximate lens choice, aspect-ratio hint, and prior score. We construct anchors from scene-bounding-box heuristics, blueprint look-toward targets, object visibility anchors, and scout-view relocation anchors. Because these anchors are decoupled from the current incumbent, they remain available when the search falls into a locally acceptable but globally weak viewpoint.

At each round, the system builds a mixed seed pool before asking the LLM to propose candidates. A seed is a partially specified camera hypothesis, usually derived from the current incumbent, a promising memory region, a global anchor, or a geometry probe. Region memory is produced by the Reflector from previous rounds: each rendered candidate is assigned to a coarse spatial cell and the cell stores visits, scores, failures, and improvement signals. Promising regions receive local refinement seeds; unknown or dead regions increase the share of global anchors and geometry probes. The LLM then turns the seed pool and reviewer feedback into complete candidate proposals
\[
  y_j = (c_j,\rho_j), \quad c_j=(p_j,\ell_j,f_j,d_j,r_j),
\]
where $\rho_j$ is a short rationale used only for interpretation and later reflection. If model output is malformed or underspecified, the implementation falls back to seed candidates and lightweight perturbations so that the loop remains executable.

\subsection{Reviewer}

The Reviewer is designed to expose why an image fails. The environment first computes rule-based indicators from projection geometry and task constraints. For example, subject visibility is estimated by projecting the target object's bounding box into the camera frame, placement and scale are measured from the projected screen box, and hard failures mark missing subjects, extreme occlusion, invalid cameras, or gross violations of required view type. A visual reviewer then scores the rendered preview along four dimensions: composition quality, technical quality, aesthetic quality, and semantic alignment. Together with the two rule-side signals, the six Reviewer signals are combined as
\begin{equation}
  J(c)=0.10m_1+0.10m_2+0.15m_3+0.15m_4+0.25m_5+0.25m_6,
  \label{eq:internal}
\end{equation}
where $m_1,m_2$ are deterministic projection-side signals and $m_3,\ldots,m_6$ are VLM-side image scores. The fixed weights are set before held-out evaluation and used only for internal search. The score is not a final evaluation metric; it ranks candidates within a run, while the dimension-wise reasoning is passed to the Reflector.

The Reviewer also performs pairwise incumbent selection. Instead of greedily replacing the best image by scalar score alone, it compares the current incumbent and the new candidate image directly, identifies the stronger image per dimension, and selects the image that is both better and more stable for subsequent optimization. This reduces oscillation when scalar scores are noisy.

\begin{table}[t]
  \caption{\textbf{Internal Reviewer signals.} The paper-level notation separates two rule-side geometric signals from four VLM image-side scores.}
  \label{tab:reviewer_signals}
  \centering
  \small
  \setlength{\tabcolsep}{3pt}
  \begin{tabular}{p{.13\linewidth}p{.16\linewidth}p{.19\linewidth}p{.40\linewidth}}
    \toprule
    Paper notation & Code field & Source & Meaning \\
    \midrule
    $m_1$ & \texttt{rule\_m1} & Blender projection & Whether the primary subject is inside the frame and in the coarse screen half requested by the composition rule. \\
    $m_2$ & \texttt{rule\_m2} & Blender projection & Normalized distance from the projected subject center to the target composition point. \\
    $m_3$ & \texttt{m1} & VLM Reviewer & Composition quality of the rendered preview. \\
    $m_4$ & \texttt{m2} & VLM Reviewer & Technical image quality, including readability and rendering artifacts. \\
    $m_5$ & \texttt{m3} & VLM Reviewer & Aesthetic quality and photographic appeal. \\
    $m_6$ & \texttt{m4} & VLM Reviewer & Alignment with the language instruction and soft blueprint. \\
    \bottomrule
  \end{tabular}
\end{table}

The Reviewer combines deterministic projection checks with VLM-based visual judgment. For the two rule-side signals, Blender projects the primary subject center into normalized screen coordinates $(u,v)$. If the subject center is outside $[0,1]\times[0,1]$, or if a left/right/top/bottom composition preference is violated at the half-screen level, $m_1=0$; otherwise $m_1=1$. The target point for $m_2$ is $(0.5,0.5)$ by default and moves to the corresponding third point for rule-of-thirds preferences. The score is $m_2=\max(0,1-d/0.45)$, where $d$ is the Euclidean screen-space distance from $(u,v)$ to the target point.

For the four VLM signals, the Reviewer receives the candidate camera parameters and the rendered preview image. It must return JSON fields \texttt{m1}, \texttt{m2}, \texttt{m3}, \texttt{m4}, \texttt{reasoning}, and \texttt{summary}. The implementation clamps each score to $[0,1]$; if parsing fails, the candidate receives a neutral fallback score of $0.5$ on all four VLM dimensions. The scalar in Eq.~\ref{eq:internal} is used only for internal ranking, region-memory updates, and search diagnostics; all main results in the paper use external post-hoc image metrics.

The Reviewer also produces structured language feedback for the next round. Given all candidate records in a round, it outputs a JSON object with \texttt{round\_review}, \texttt{next\_strategy}, \texttt{step\_scale}, \texttt{explore\_ratio\_next}, \texttt{preferred\_motion}, \texttt{failure\_tags}, \texttt{forbidden\_zones}, and optional seed \texttt{candidates}. The implementation clamps \texttt{step\_scale} to $[0.4,1.8]$, clamps \texttt{explore\_ratio\_next} to $[0.1,0.8]$, keeps at most six failure tags, accepts at most two Reviewer-generated forbidden zones, normalizes candidate camera parameters, and merges Reviewer forbidden zones with Reflector dead regions. Pairwise incumbent selection is handled separately: each new preview is compared against the current incumbent, and a parsing failure falls back to keeping the incumbent. These constraints make the Reviewer a schema-bounded search controller rather than an unconstrained conversational critic.

We considered preference-based Bayesian optimization (PBO) for this selection step because recent agentic aerial cinematography uses pairwise visual preferences to refine 6-DoF camera poses \citep{agenticAerial2025}. In that setting, however, the optimizer repeatedly samples many challenger poses per update (e.g., 64 candidates per iteration) and may require tens to roughly one hundred preference updates before convergence. Such sampling is expensive for virtual photography, where every pose comparison requires rendering a candidate image. In our tests, frontier vision-language models already produced pairwise image comparisons close to human preference for near-neighbor photographic choices, so \methodname{} uses direct reviewer comparison to choose the round incumbent instead of running a separate PBO loop.

\subsection{Reflector}

The Reflector turns round-level feedback into future control signals. Continuous space is discretized into cubic region cells with side length $h=\max(0.12\,\mathrm{sceneScale},0.9)$. Each region records visit count, best score, semantic score, poor hits, promising hits, improvement hits, and stagnation hits, and is labeled as unknown, promising, or dead. A region becomes promising if its best internal score reaches $.68$, its best semantic score reaches $.70$, or it receives a promising hit; it becomes dead after repeated low-score visits or repeated stagnation without improvement. Dead regions are converted into forbidden zones, while promising regions may still be exploited.

To prevent premature local collapse, the architecture includes a forced high-explore lane. In each round, when feasible, one anchor seed $a$ is drawn from the global anchor bank according to a priority score
\begin{equation}
  s(a) =
  \pi(a) + u(a) +
  \min\!\left(\frac{\|p_a-p_t\|_2}{2h}, 2.0\right)
  -0.35\, n(a)
  -0.40\,\mathbf{1}[a \in \mathrm{promising}],
  \label{eq:highexplore}
\end{equation}
where $\pi(a)$ is the anchor prior from scouting, $p_a$ and $p_t$ are the anchor and current-incumbent camera positions, $n(a)$ is the visit count of the anchor's region, and $u(a)=1.2$ for an unknown region and $.25$ otherwise. Anchors in dead regions are skipped before ranking. This is not random restart. It is a structured curiosity channel that keeps one candidate exploring low-visit, non-dead, spatially meaningful anchors, optionally with a different aspect ratio.

\subsection{Rendering and framing}

Rendering is the main systems bottleneck in iterative virtual photography. \methodname{} decouples candidate preview rendering from agent logic by launching external Blender subprocesses when the source scene and binary path permit it. Preview samples are capped at 64 and render settings are restored afterward, so final render quality is not polluted by preview settings. If parallel preview is unavailable, the implementation falls back to serial rendering. Run logs store preview caps, worker counts, final samples, selected aspect ratio, image paths, and model/backend options; the public release will include the prompt templates, JSON schemas, run configurations, and evaluation scripts used to reproduce the paper.

Aspect ratio is also handled as a compositional decision. Candidate proposals must choose an aspect ratio from $A$ and justify it in the candidate rationale. After search, the system reruns a final aspect-ratio selection step using the best preview image, scene axis strength, subject concentration, environmental breadth, and requested atmosphere. The final output is rendered at a resolution derived from the selected ratio.

\section{\benchmarkname: Benchmark Formulation}

\subsection{Benchmark composition}
\label{sec:benchmark_composition}

\benchmarkname{} instantiates the task formulation from Section~\ref{sec:task_formulation} over 47 open-license Blender scenes. 28 scenes come from the official Blender Demo Files archive~\citep{blender2025}, and 19 come from Blend Swap~\citep{blendswap2025}. Each scene is paired with three natural-language missions---subject placement, relational composition, and atmosphere/style---yielding 141 runnable task instances. Table~\ref{tab:benchmark} reports the scene distribution over visual style, environment, and subject type. Each scene also receives a five-level complexity rating: annotators manually inspect the scene layout and assign a one-to-five star rating as an auxiliary indicator of spatial and compositional difficulty. The release package will include the scene registry, task JSON files, evaluation specifications, and per-scene source/license metadata; original assets remain governed by their upstream licenses.

\begin{table}[t]
  \caption{\textbf{Benchmark composition.} We summarize the same set of 47 scenes along three diversity axes. Each scene contains three tasks, yielding 141 tasks in total.}
  \label{tab:benchmark}
  \centering
  \small
  \setlength{\tabcolsep}{4pt}
  \begin{tabular}{llcc}
    \toprule
    Category & Subcategory & \#Scenes & Total tasks \\
    \midrule
    \multirow{4}{*}{Visual style}
      & Stylized / Cartoon & 16 & 48 \\
      & Realistic & 15 & 45 \\
      & Fantasy / Mystical & 9 & 27 \\
      & Sci-Fi / Cyberpunk & 7 & 21 \\
    \midrule
    \multirow{4}{*}{Environment}
      & Outdoor / Natural & 18 & 54 \\
      & Indoor / Interior & 12 & 36 \\
      & Abstract / Mixed & 12 & 36 \\
      & Space / Cosmic & 5 & 15 \\
    \midrule
    \multirow{4}{*}{Subject type}
      & Panoramic / No hero & 19 & 57 \\
      & Architecture & 15 & 45 \\
      & Nature / Object & 7 & 21 \\
      & Character / Creature & 6 & 18 \\
    \midrule
    \multicolumn{2}{l}{Benchmark total} & 47 & 141 \\
    \bottomrule
  \end{tabular}
\end{table}

\begin{figure}[t]
  \centering
  \includegraphics[width=.88\linewidth]{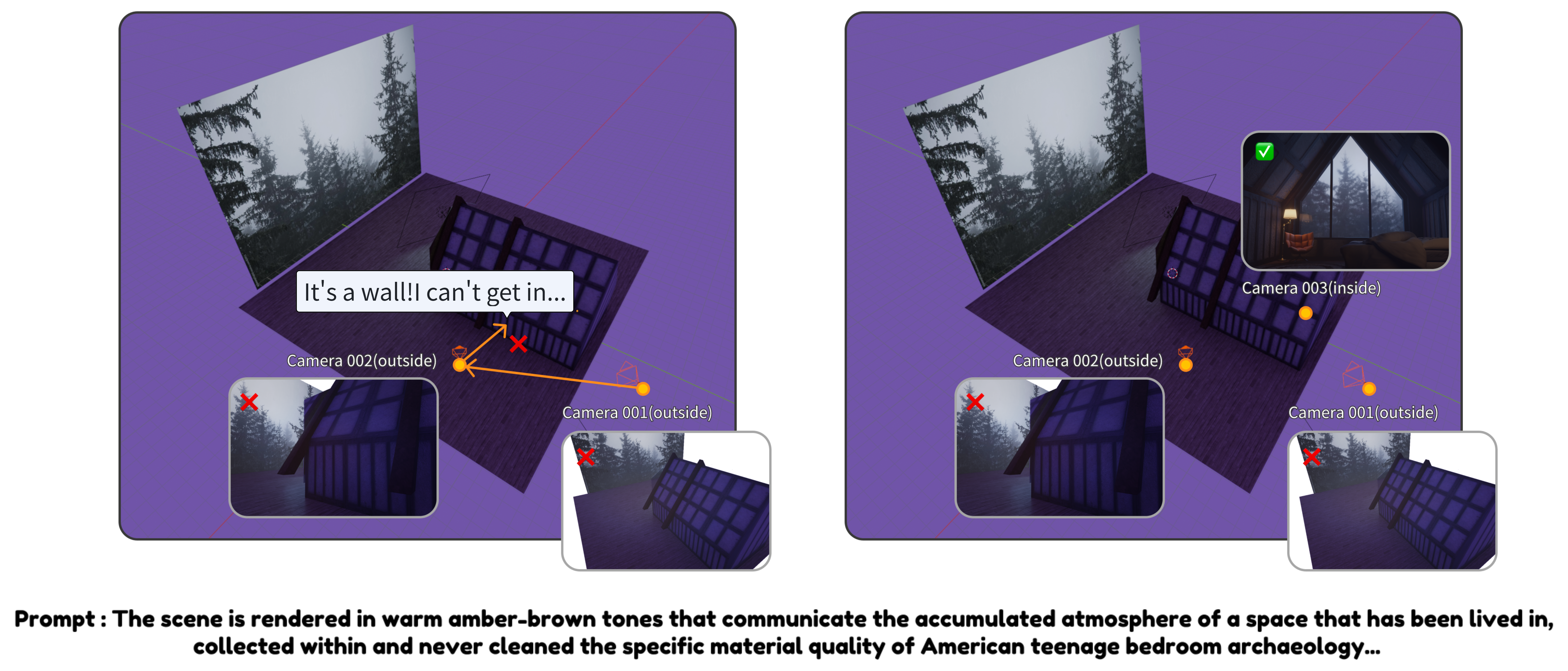}
  \caption{\textbf{Task boundary with VLN.} VLN is a useful neighboring formulation because both settings make language-conditioned 3D decisions, but VLN evaluates navigation paths while virtual photography evaluates the final executable camera state and rendered view.}
  \label{fig:vln_boundary}
\end{figure}

\begin{figure}[t]
  \centering
  \includegraphics[width=.78\linewidth]{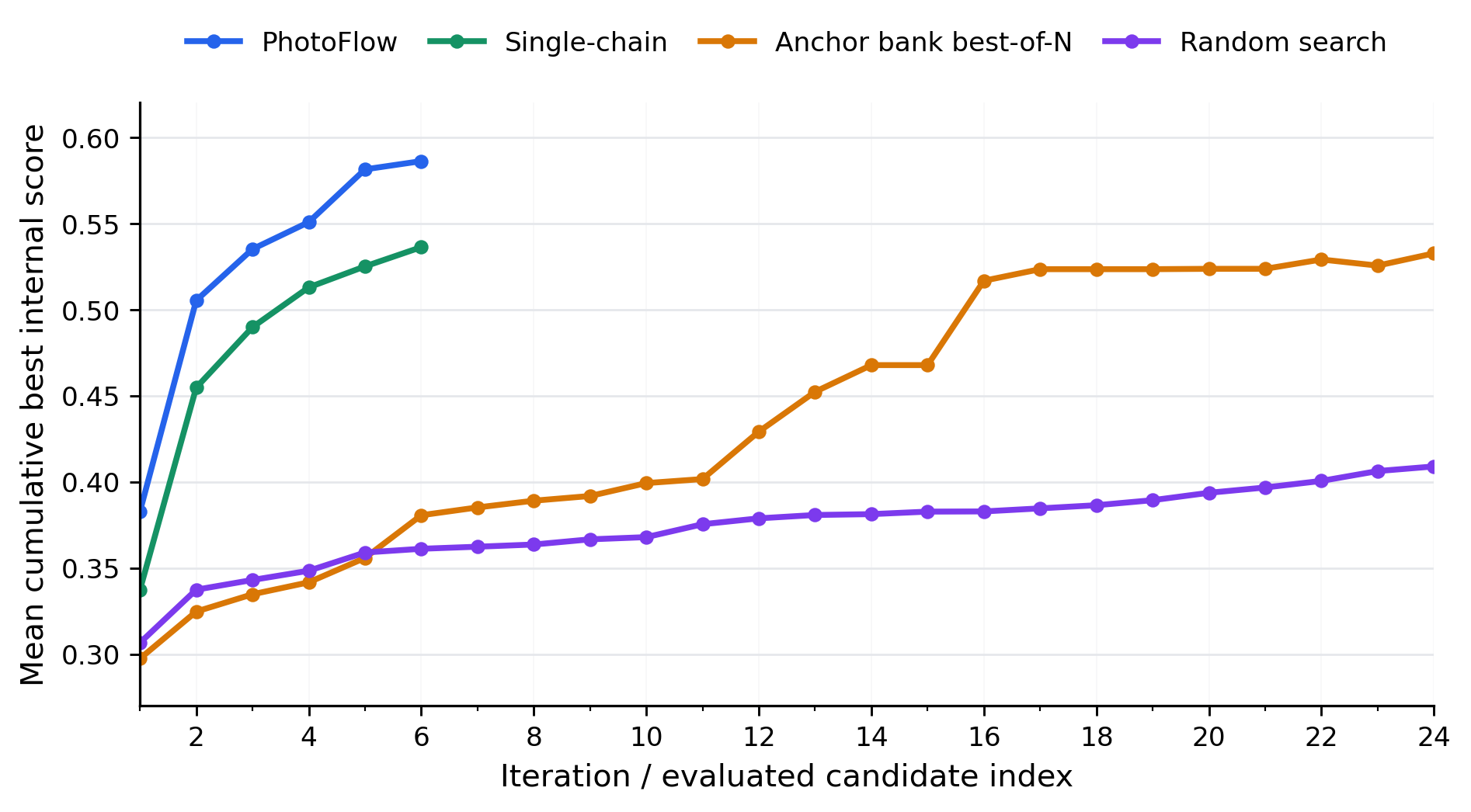}
  \caption{\textbf{Search-process diagnostic.} Internal cumulative best score during search. The horizontal axis is feedback round for iterative methods and evaluated-candidate index for one-shot candidate pools. External image metrics in Table~\ref{tab:main} remain the main evidence.}
  \label{fig:search_diagnostic}
\end{figure}

\section{Experiments}

We evaluate \methodname{} by separating two questions: whether the benchmark exposes spatial-aesthetic failures that are invisible to single-score evaluation, and whether a closed-loop Director-Reviewer-Reflector search improves camera selection under a fixed rendering budget. Because the agent uses an internal Reviewer during optimization, final comparisons are based on external image metrics and human consistency checks rather than internal scores alone; constraint logs are retained only for failure accounting and diagnosis.

\subsection{Protocol}

The main benchmark unit is $(\mathrm{scene}, \mathrm{instruction})$. We use 24 development missions for prompt and threshold selection and reserve 117 missions for held-out testing. Each method is launched on the same 117 held-out missions with matched final render settings, external evaluators, and random seeds. For image-quality means, we apply a task-level common-completed rule: a task is included only if every compared method produces a final image and external scores; otherwise it remains in the failure log. This leaves 90 common completed tasks and excludes the same 27 task IDs for every method because their scenes triggered systems-level render failures under large-scale multi-method evaluation: 21 no-first-image timeouts, 3 no-final-image events, and 3 Blender crashes per method. The rule is method-independent rather than winner-selective: retained tasks preserve an exactly balanced mission split of 30 subject-placement, 30 relational-composition, and 30 atmosphere/style missions, and cover all scene families and five complexity levels. This filter is necessary because Blender render time is scene-dependent, with some authored scenes requiring hours for a single final image, but it does not change the 141-task benchmark definition.

For iterative methods, the main protocol uses a low search budget of $T=6$ rounds and four preview candidates per round for \methodname{}; completed runs average 20.8 preview renders because malformed proposals, retries, or unavailable preview slots can reduce the realized count. Random Search evaluates 24 independent candidate views, Iterative Single-Chain Reflection evaluates one preview per round for six rounds, Anchor Bank Best-of-$N$ scans the generated scout/anchor bank (mean 12.6 anchors, median 11), and Single-Step LLM renders one final prediction. These controls are designed to isolate sources of performance rather than pretend that all baselines have identical agent structure. We do not impose an additional hand-tuned early-stop threshold; within the fixed budget, the agent selects its final incumbent through reviewer comparison and reflection. All final images are scored only after the search procedure completes.

The primary comparison includes Single-Step LLM, Iterative Single-Chain Reflection, Anchor Bank Best-of-$N$, Random Search, and \methodname{}. Anchor Bank Best-of-$N$ is a critical strong baseline: it uses the same scout and anchor bank as \methodname{} but removes reflection and cross-round memory, testing whether gains come from the agent loop rather than good initial anchors alone.

\paragraph{Baseline scope.}
VLN was the closest embodied formulation we considered during baseline design because it also studies language-conditioned decisions in 3D environments. This comparison clarified the boundary of the new task (Figure~\ref{fig:vln_boundary}). VLN assumes a navigable graph and evaluates a policy over movement, with the final stop and the path both contributing to success \citep{anderson2018r2r,anderson2018evaluation,jain2019stay}. Virtual photography instead evaluates the final executable camera state: once a camera pose, look-at point, lens, aperture, and aspect ratio are selected, the path used to discover them has no effect on the rendered image. This distinction matters in authored virtual scenes, where geometry is often arranged for appearance rather than physical traversability. We therefore compare against baselines that directly optimize final camera selection, while using VLN as related evidence that language-conditioned 3D decision making is a natural but not identical neighboring problem.

\paragraph{Metrics.}
We use external metrics for all main comparisons. UniPercept \citep{cao2025unipercept} provides three image-side scores: image aesthetic assessment $M_{\mathrm{iaa}}$, image quality assessment $M_{\mathrm{iqa}}$, and image structure-texture/alignment assessment $M_{\mathrm{ista}}$. The primary quality-alignment score is
\begin{equation}
  M_{\mathrm{qs}} =
  \omega_a M_{\mathrm{iaa}} +
  \omega_q M_{\mathrm{iqa}} +
  \omega_s M_{\mathrm{ista}},
  \quad
  (\omega_a,\omega_q,\omega_s)=(0.40,0.20,0.40).
  \label{eq:qualitysemantic}
\end{equation}
We choose a larger weight for aesthetics and ISTA because a virtual photograph must both look strong and preserve the requested visual structure and style; technical quality is still included, but has lower weight because all methods use the same renderer. We additionally report $\mathrm{Succ@0.55}$, the fraction of tasks whose $M_{\mathrm{qs}} \ge 0.55$, as a thresholded quality-success measure. Structured constraint logs and hard-failure tags are retained for failure accounting and qualitative diagnosis, but we do not use them as main ranking columns because the current export is log-based rather than a Blender ray-cast recomputation and is therefore less architecture-neutral than the external image metrics.

\begin{table}[t]
  \caption{\textbf{Main comparison on 90 common completed held-out tasks.} $M_{\mathrm{qs}}$ is the external quality-alignment composite in Eq.~\ref{eq:qualitysemantic}. $\mathrm{Succ@0.55}$ is the fraction of tasks with $M_{\mathrm{qs}}\ge0.55$.}
  \label{tab:main}
  \centering
  \small
  \setlength{\tabcolsep}{4pt}
  \begin{tabular}{lccccc}
    \toprule
    Method & $M_{\mathrm{qs}} \uparrow$ & Succ@0.55 $\uparrow$ & IAA $\uparrow$ & IQA $\uparrow$ & ISTA $\uparrow$ \\
    \midrule
    Single-Step LLM & $.514{\pm}.130$ & $.400$ & $.447$ & $.470$ & $.603$ \\
    Anchor Bank Best-of-$N$ & $.519{\pm}.110$ & $.378$ & $.464$ & $.481$ & $.593$ \\
    Random Search & $.527{\pm}.124$ & $.400$ & $.483$ & $.492$ & $.589$ \\
    Single-Chain Reflection & $.567{\pm}.115$ & $.567$ & $.530$ & $.545$ & $\mathbf{.616}$ \\
    \textbf{\methodname{}} & $\mathbf{.578{\pm}.120}$ & $\mathbf{.622}$ & $\mathbf{.550}$ & $\mathbf{.564}$ & $.614$ \\
    \bottomrule
  \end{tabular}
\end{table}

Table~\ref{tab:main} shows that the closed-loop agent improves the primary external quality-alignment composite over all tested baselines under a six-round budget. The gain is largest over one-shot and anchor-only policies, supporting the claim that feedback-driven search adds value beyond strong initial viewpoint priors; at the per-task level, \methodname{} wins 68/90 tasks against Anchor Bank Best-of-$N$ and 60/90 against Random Search. The strongest baseline is Iterative Single-Chain Reflection, which slightly leads ISTA but trails \methodname{} on the combined score, success rate, aesthetics, and image quality; this comparison is close, with \methodname{} winning 49/90 paired tasks, so we interpret the gain as modest rather than overwhelming. This pattern matches the goal of the benchmark: the main question is not whether every diagnostic scalar is maximized, but whether the agent can produce stronger final photographs when spatial intent and aesthetic judgment must be solved together.

\begin{table}[t]
  \caption{\textbf{Per-category results.} This table tests whether improvements hold across mission types instead of being dominated by a single scene family.}
  \label{tab:category}
  \centering
  \small
  \begin{tabular}{lccc}
    \toprule
    Method & Subject placement & Relational composition & Atmosphere/style \\
    \midrule
    Single-Step LLM & $.499$ & $.526$ & $.517$ \\
    Anchor Bank Best-of-$N$ & $.510$ & $.525$ & $.522$ \\
    Random Search & $.514$ & $.523$ & $.544$ \\
    Single-Chain Reflection & $.560$ & $.577$ & $.565$ \\
    \textbf{\methodname{}} & $\mathbf{.578}$ & $\mathbf{.582}$ & $\mathbf{.574}$ \\
    \bottomrule
  \end{tabular}
\end{table}

\paragraph{Search process.}
Figure~\ref{fig:search_diagnostic} plots the internal cumulative best score during search. This curve is not used as final evidence, because the internal Reviewer is part of the agent, but it explains how the methods behave before external evaluation. \methodname{} reaches a high internal incumbent within six rounds, while one-shot pools improve more slowly as more candidates are evaluated. The process evidence supports the interpretation that the Director-Reviewer-Reflector loop is doing structured search, not merely relying on a single lucky anchor.

\subsection{Ablations}

The ablation study is organized around the three-role design. The Director cannot be removed wholesale because every runnable method must still propose executable camera poses; instead, its multi-candidate behavior is tested by the single-chain baseline, which removes parallel preview rendering and keeps only one feedback trajectory. The Reviewer role is tested by the anchor-bank best-of-$N$ baseline, which removes structured review and reflection while preserving the same initial anchor pool. Table~\ref{tab:ablation} therefore focuses on the remaining removable mechanisms: region memory for the Reflector, and high-explore relocation as the special safeguard against local collapse. In addition to external quality, we report raw-log search diagnostics: region coverage, local collapse, and revisit rate. These diagnostics are not geometric correctness metrics; they explain whether a variant searches too narrowly, revisits low-value regions, or loses exploration diversity.

\begin{table}[t]
  \caption{\textbf{Ablation study and search diagnostics.} Image metrics are external scores on the same 90 common completed tasks. Coverage, Collapse, and Revisit are log-based search diagnostics computed from candidate region keys.}
  \label{tab:ablation}
  \centering
  \scriptsize
  \setlength{\tabcolsep}{3pt}
  \resizebox{\linewidth}{!}{%
  \begin{tabular}{lcccccccc}
    \toprule
    Variant & $M_{\mathrm{qs}} \uparrow$ & Succ@0.55 $\uparrow$ & IAA $\uparrow$ & IQA $\uparrow$ & ISTA $\uparrow$ & Coverage $\uparrow$ & Collapse $\downarrow$ & Revisit $\downarrow$ \\
    \midrule
    Full \methodname{} & $.578$ & $\mathbf{.622}$ & $.550$ & $.564$ & $.614$ & $\mathbf{.453}$ & $\mathbf{.389}$ & $\mathbf{.547}$ \\
    w/o region memory & $.572$ & $.544$ & $.539$ & $.559$ & $.611$ & $.356$ & $.411$ & $.644$ \\
    w/o high-explore & $\mathbf{.584}$ & $.611$ & $\mathbf{.557}$ & $\mathbf{.576}$ & $\mathbf{.616}$ & $.331$ & $.456$ & $.669$ \\
    \bottomrule
  \end{tabular}
  }
\end{table}

\begin{figure}[t]
  \centering
  \includegraphics[width=.82\linewidth]{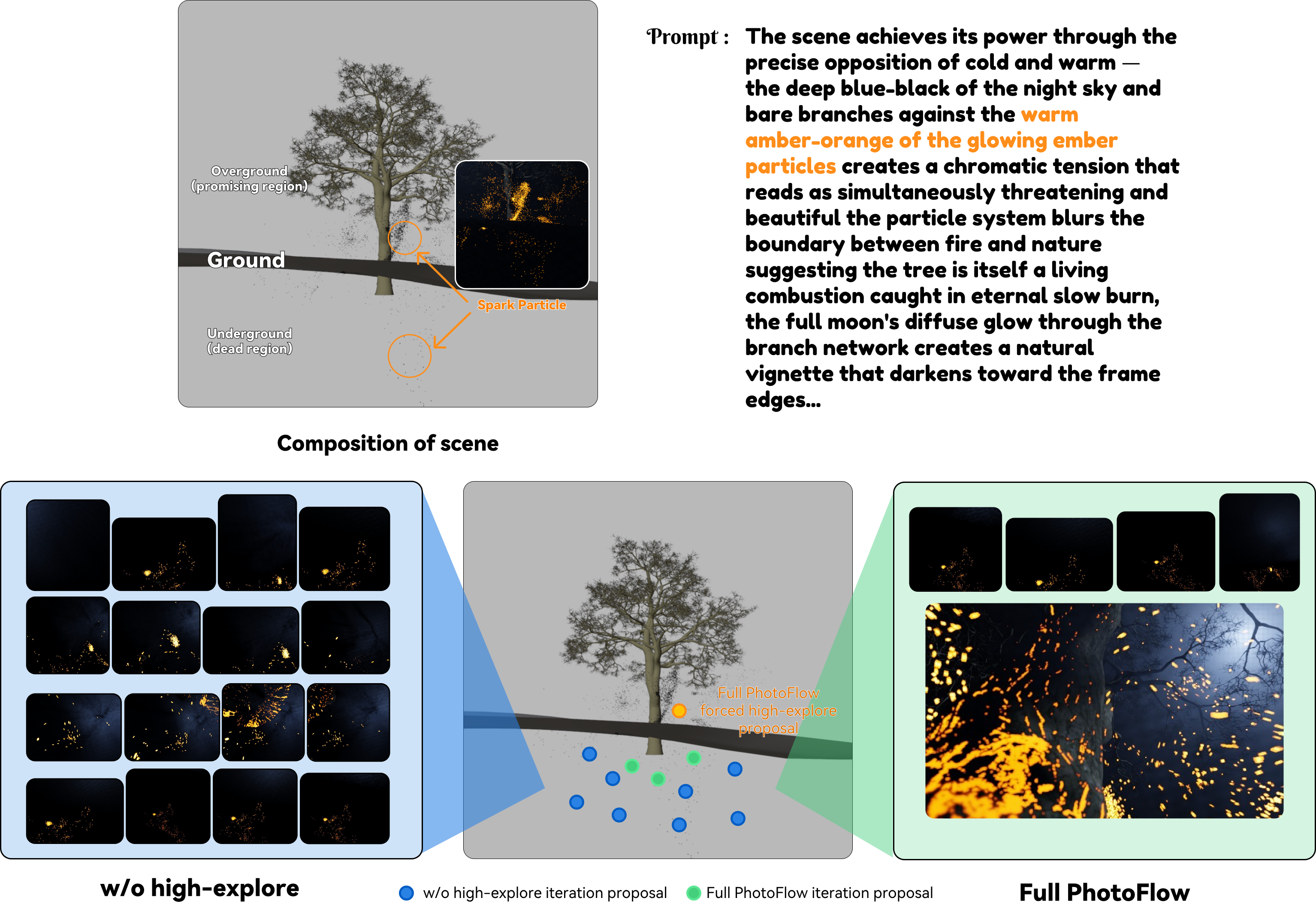}
  \caption{\textbf{High-explore as a switchable safeguard.} A representative case where forced high-explore helps the search leave a locally acceptable but weak viewpoint and find a stronger composition. This component is intended as an escape route from local collapse, not as a universally better proposal source.}
  \label{fig:high_explore_case}
\end{figure}

Table~\ref{tab:ablation} shows that the component effects are not purely monotonic, which is expected for a finite-budget search agent. Region memory provides the cleanest support for the Reflector design: removing it lowers external quality and success while increasing revisits. High-explore relocation has a different role. Disabling it raises some external averages on this subset, but also substantially lowers coverage and increases collapse/revisit rates, matching its intended use as a switchable safeguard rather than a universally beneficial proposal source. We also tested stricter heuristic stopping during development, but it did not provide a reliable quality benefit; the final system therefore keeps the simple six-round cap and lets the agent choose the final incumbent instead of adding another hard stop rule.

High-explore deserves separate interpretation because it is a safeguard rather than a universally beneficial proposal source. It was introduced to escape the local-optimum failures observed during development, and the raw logs support this role: for example, in the forest subject-placement task, the full system improves $M_{\mathrm{qs}}$ from $.527$ without high-explore to $.696$. At the same time, a forced high-explore lane consumes one candidate slot, so it can reduce the number of direct refinement candidates and make results less stable on some scenes. We therefore expose it as a switchable component and include a qualitative case in Figure~\ref{fig:high_explore_case}.

\subsection{Human consistency}

Because visual quality cannot be fully reduced to automatic metrics, we run a two-stage human subset study. The cleaned export keeps the latest duplicate response per participant/question and retains participants who completed the full survey; 30 of 31 participants pass this quality-control rule, yielding 780 valid response rows, including 450 multi-image preference responses and 300 PhotoFlow-only Likert ratings. The purpose is consistency checking, not replacing the main benchmark: Stage~1 asks which image is preferred for aesthetics and instruction alignment among the compared methods, while Stage~2 asks for mean-opinion scores on PhotoFlow outputs and correlates them with automatic metrics.

The full results in Table~\ref{tab:human} support the automatic evaluation as an imperfect but informative diagnostic. \methodname{} receives the highest selection rate in both aesthetic and alignment choices, with Iterative Single-Chain Reflection as the closest non-ours method; the recorded 95\% intervals for PhotoFlow selection are $.271$--$.356$ for aesthetics and $.260$--$.347$ for alignment. $M_{\mathrm{qs}}$ also correlates strongly with human MOS mean on the rated subset. These rates are not large enough to claim overwhelming human preference, but they support $M_{\mathrm{qs}}$ as a practical main metric while leaving room for human disagreement and future evaluator improvement.

\begin{table}[t]
  \caption{\textbf{Human consistency study.} Stage~1 reports selection rates; Stage~2 reports PhotoFlow-only MOS and correlation with automatic metrics.}
  \label{tab:human}
  \centering
  \small
  \resizebox{\linewidth}{!}{%
  \begin{tabular}{lcccc}
    \toprule
    Evaluation & Responses / points & \textbf{PhotoFlow} & Strongest non-ours & Additional statistic \\
    \midrule
    Aesthetic selection & 450 & $\mathbf{31.33\%}$ & $24.89\%$ single-chain & ties $12.67\%$ \\
    Alignment selection & 450 & $\mathbf{30.22\%}$ & $23.11\%$ single-chain & ties $15.33\%$ \\
    PhotoFlow MOS & 300 & $\mathbf{3.208}$ mean & -- & aesthetic $3.313$, alignment $3.103$ \\
    $M_{\mathrm{qs}}$ vs. MOS mean & 24 & Pearson $.827$ & Spearman $.697$ & PhotoFlow-only ratings \\
    \bottomrule
  \end{tabular}
  }
\end{table}

\subsection{Qualitative case studies}
\label{sec:qualitative_cases}

\begin{figure}[t]
  \centering
  \includegraphics[width=.96\linewidth]{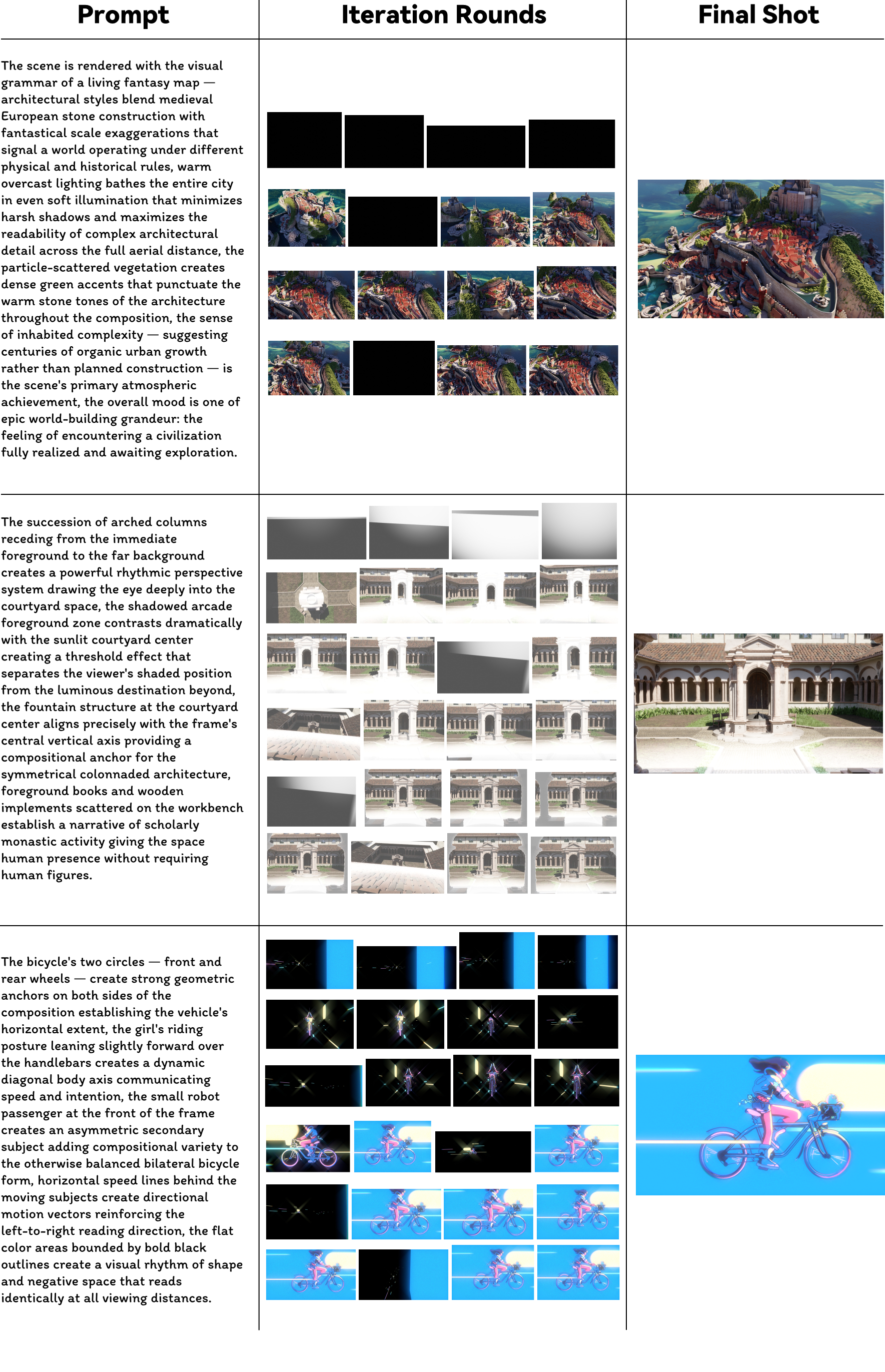}
  \caption{\textbf{Successful qualitative cases.} Each row is organized as prompt, iterative previews, and final render. The three examples cover a city/island composition, a courtyard architecture view, and a stylized bicycle subject, showing how \methodname{} turns language into a sequence of rendered camera hypotheses and a final executable camera state across different scene scales and visual styles.}
  \label{fig:success_cases}
\end{figure}

\begin{figure}[t]
  \centering
  \includegraphics[width=.96\linewidth]{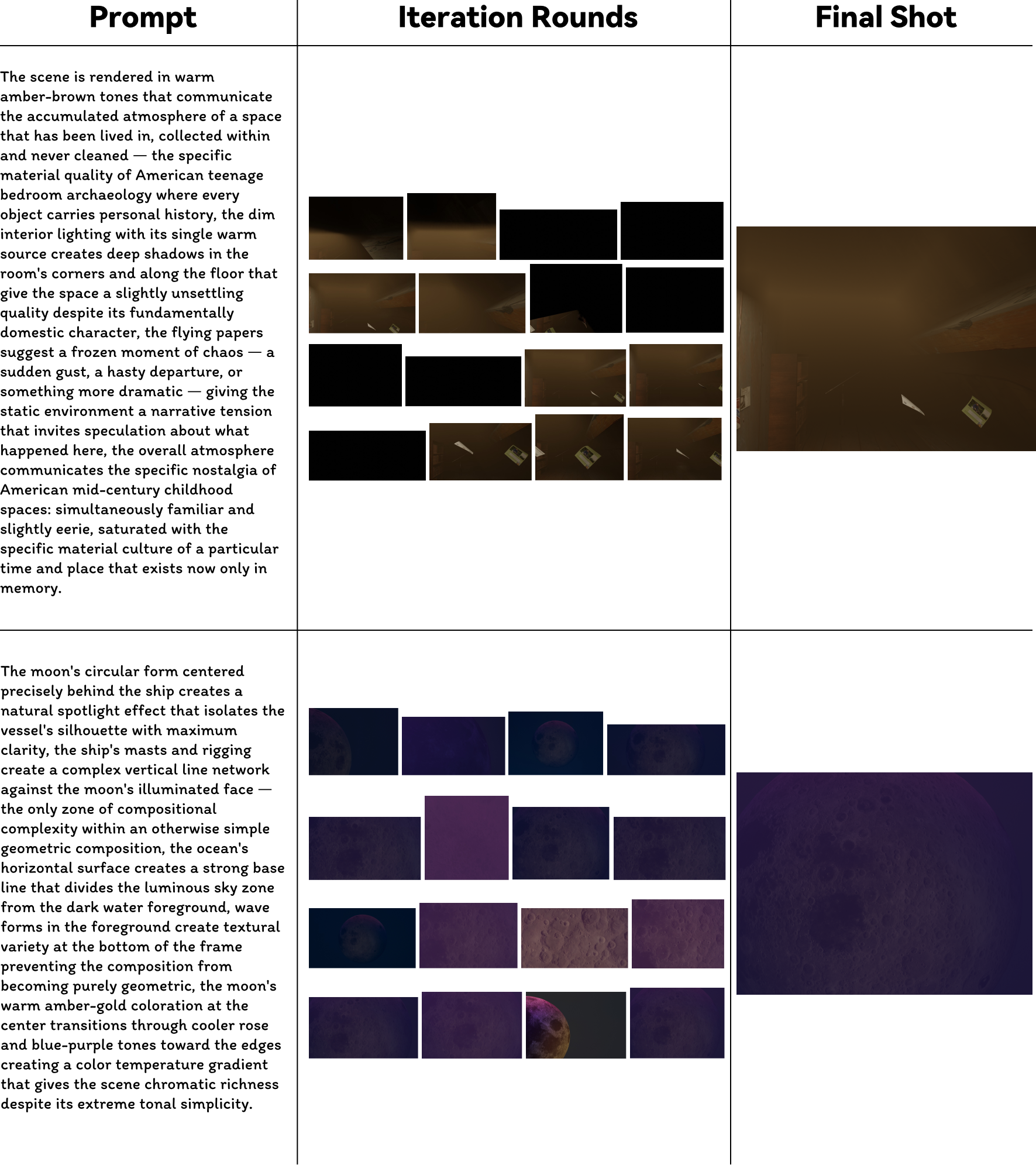}
  \caption{\textbf{Failure qualitative cases.} Each row is organized as prompt, iterative previews, and final render. The top row is `037\_attic\_hideout\_atmosphere\_style': the search collapses into a dark, low-quality atmospheric view and receives a hard-failure tag with constraint satisfaction $0.0$ ($M_{\mathrm{qs}}=.244$). The bottom row is `031\_medieval\_ship\_ocean\_scene\_subject\_placement': the final camera fails the requested subject/framing constraint despite partial semantic alignment, also yielding constraint satisfaction $0.0$ ($M_{\mathrm{qs}}=.338$).}
  \label{fig:failure_cases}
\end{figure}

Figures~\ref{fig:success_cases} and~\ref{fig:failure_cases} show qualitative cases in the same layout: the left column gives the language prompt, the middle column shows the iterative search previews, and the right column shows the final selected render.

\section{Limitations}

\methodname{} depends on assumptions that should be tested rather than hidden. First, the quality of global exploration is bounded by the anchor bank; if the scene scout and visibility anchors miss the relevant region, the high-explore lane may still be weak. Second, the Reviewer supplies useful diagnostic feedback, but its internal scores are not sufficient as final evidence; the main protocol therefore requires external evaluators and human preference checks. Third, the reported main table is a common-completed image-quality comparison, not an end-to-end availability score over all render-heavy scenes; future releases should make failure logs, timeout categories, and render-time buckets auditable. Fourth, the current study does not include formal threshold-sensitivity plots, paired significance tests, or adapted PBO/geometric-planner baselines, so the closest improvements should be read as measured evidence for this benchmark rather than as a universal dominance claim. Finally, virtual photography is a controlled proxy for creative camera work. The benchmark improves reproducibility, but results may not transfer directly to physical robots or dynamic scenes without additional control, collision, and temporal-consistency constraints.

\section{Conclusion}

We presented \benchmarkname{} and \methodname{} as a benchmark-and-agent framework for language-conditioned virtual photography. The benchmark turns aesthetic camera selection into a reproducible task-level protocol, and the agent turns photography into structured closed-loop search over executable camera states. The central claim is deliberately narrow: virtual photography needs evaluation that jointly measures spatial constraints and aesthetic intent, and camera-search agents need structured feedback to diagnose and escape poor local viewpoints. This framing provides a concrete path for measuring spatial-aesthetic intelligence in controllable 3D worlds.

\bibliographystyle{plainnat}
\bibliography{references}

\end{document}